\documentclass[11pt]{article}
\usepackage{enumerate}
\usepackage{pdfsync}
\usepackage[OT1]{fontenc}

\usepackage[usenames]{color}
\usepackage{smile}
\usepackage[colorlinks,
            linkcolor=red,
            anchorcolor=blue,
            citecolor=blue
            ]{hyperref}
\usepackage{fullpage}
\usepackage[protrusion=true,expansion=true]{microtype}
\usepackage{pbox}
\usepackage{setspace}
\usepackage{tabularx}
\usepackage{float}
\usepackage{wrapfig,lipsum}
\usepackage{enumitem}
\usepackage{microtype}
\usepackage{graphicx}
\usepackage{subfigure}
\usepackage{pgfplots}
\usepackage{booktabs} 

\allowdisplaybreaks
\usepackage{colortbl}

\usepackage{xcolor}

\makeatletter
\newcommand*{\rom}[1]{\expandafter\@slowromancap\romannumeral #1@}
\makeatother
\title{\huge Hallucination Detection and Evaluation of Large Language Model}
\author
{
    Chenggong Zhang\thanks{Department of Electrical and Computer Engineering, University of California, Los Angeles, CA 90095, USA; email: {\tt chenggong61@g.ucla.edu}}~~~
    Haopeng Wang \thanks{Department of Electrical and Computer Engineering, University of California, Los Angeles, CA 90095, USA; email: {\tt hpwang@g.ucla.edu}}~~~
    Hexi Meng \thanks{Department of Electrical and Computer Engineering, University of California, Los Angeles, CA 90095, USA; email: {\tt fury1120@g.ucla.edu}}
}

\begin{document}
    \date{}
    \maketitle

\begin{abstract}
Hallucinations in Large Language Models (LLMs) pose a significant challenge, generating misleading or unverifiable content that undermines trust and reliability. Existing evaluation methods, such as KnowHalu, employ multi-stage verification but suffer from high computational costs. To address this, we integrate the Hughes Hallucination Evaluation Model (HHEM), a lightweight classification-based framework that operates independently of LLM-based judgments, significantly improving efficiency while maintaining high detection accuracy. We conduct a comparative analysis of hallucination detection methods across various LLMs, evaluating True Positive Rate (TPR), True Negative Rate (TNR), and Accuracy on question-answering (QA) and summarization tasks. Our results show that HHEM reduces evaluation time from 8 hours to 10 minutes, while HHEM with non-fabrication checking achieves the highest accuracy \(82.2\%\) and TPR \(78.9\%\). However, HHEM struggles with localized hallucinations in summarization tasks. To address this, we introduce segment-based retrieval, improving detection by verifying smaller text components. Additionally, our cumulative distribution function (CDF) analysis indicates that larger models (7B-9B parameters) generally exhibit fewer hallucinations, while intermediate-sized models show higher instability. These findings highlight the need for structured evaluation frameworks that balance computational efficiency with robust factual validation, enhancing the reliability of LLM-generated content.
\end{abstract}

\section{Introduction}

Hallucinations with AI are part of a growing list of ethical concerns with respect to the LLM model. Aside from misleading people with factually inaccurate information and eroding user trust, hallucinations can perpetuate biases or cause other harmful consequences if taken at face value. Therefore, detecting and reducing hallucinations precisely becomes a serious problem.

For this project, our aim is to improve the evaluation framework effectiveness through integration and compare and analyze the performance differences of models of different scales in the hallucination detection task using different evaluations and develop effective structure to reduce hallucination. Regarding the hallucination detection and evaluation, researchers have developed and implemented a hallucination called \textbf{KnowHalu} \cite{zhang2024knowhalu} To detect hallucinations in LLM models, we initially employed the KnowHalu method. However, this method has certain limitations. For example, a multistage approach time consuming and computationally heavy as it systematically processes and verifies information through each phase. To address these issues and further improve hallucination detection and evaluation, we applied an updated model
{\textbf{HHEM}} to improve the overall detection and evaluation performance. \cite{zapier_hallucinations},which computes a Factual Consistency Score to assess the probability that one text is factually consistent with another, thereby enabling faster evaluation times.

\section{Related Work} 
Besides our framework of hallucination detection and evaluation,Recent studies have explored a multi-pronged approach to mitigating hallucinations in large language models (LLMs) by combining retrieval-augmented generation (RAG), advanced prompting techniques, and fine-tuning. Retrieval-augmented generation methods first retrieve relevant, verifiable data from external sources and append it to the model’s input. This strategy has been shown to significantly reduce the incidence of hallucinated outputs by ensuring that the generated responses are anchored in factual information.\cite{kirkovska2024reduce}
Another notable work that leverages reinforcement learning for hallucination detection is detailed in a comprehensive survey by Towhidul Islam Tonmoy et al.\cite{tonmoy2024}, which discusses reinforcement learning techniques as part of a broader set of methods for mitigating hallucinations in LLMs. In this survey, reinforcement learning is highlighted as a means to fine-tune the model’s behavior by incorporating human feedback to adjust its confidence in generated outputs and penalize hallucinated content. Additionally, the "Chain-of-Verification" approach by Dhuliawala et al\cite{dhuliawala2023chain}. employs a multi-step process—where the model first generates a draft response and then verifies it through additional reasoning steps, a process that can be optimized using reinforcement learning methods to further reduce hallucinations.

HHEM \cite{mendelevitch2024hhem} is the first industry-standard hallucination evaluation model, introduced at the end of 2023, demonstrating state-of-the-art performance in hallucination detection. Unlike KnowHalu, HHEM is a pure classification model that does not rely on an LLM for judgment, offering significant advantages in efficiency and accessibility. It surpasses GPT-4 in performance while maintaining a compact model size of 439MB, allowing it to run efficiently on consumer-grade hardware such as an RTX 3090 GPU. At 32-bit precision, it occupies less than 600MB of RAM, making it highly memory-efficient. Additionally, on modern x86 CPUs, it processes 2K token inputs in approximately 1.5 seconds, ensuring rapid inference for hallucination detection.

HHEM evaluates hallucinations based on four key metrics: \textit{Hallucination Rate}, which measures the percentage of summaries with a hallucination score below 0.5; \textit{Factual Consistency Rate}, which is the complement of the hallucination rate and reflects the proportion of factually accurate summaries; \textit{Answer Rate}, which accounts for the percentage of non-empty summaries, including cases where the model refuses to generate a response or encounters an error; and \textit{Average Summary Length}, which tracks the average word count of generated summaries. These evaluation criteria ensure that HHEM provides a comprehensive, scalable, and efficient assessment of hallucination tendencies across different language models.

By leveraging both hallucination evaluation framework, our approach not only addresses the long evaluation times associated with the KnowHalu framework—improving its overall efficiency and performance—but also targets the low recall issue of HHEM through strategic integration.
In this study, we evaluate the performance of a diverse set of large language models by measuring their true positive rate (TPR), true negative rate (TNR), and overall accuracy. Our analysis includes Llama-based models (Llama-2-7b-chat-hf and Llama-3.1-8B-Instruct), Qwen2.5 models (Qwen2.5-0.5B-Instruct, Qwen2.5-1.5B-Instruct, Qwen2.5-3B-Instruct, and Qwen2.5-7B-Instruct), DeepSeek-R1-Distill-Llama-8B, Starling-LM-7B-alpha (baseline), gemma-7b-it, Mistral-7B-Instruct-v0.3, and internlm3-8b-instruct. This comprehensive evaluation is designed to assess various aspects of LLM hallucinations, following an approach similar to the HHEM leaderboard.

\section{Methodology}
Hallucinations in Large Language Models (LLMs) occur during inference, particularly in the post-training phase, when the model generates inaccurate or unverifiable content due to overgeneralization, lack of factual grounding, or reliance on incomplete knowledge. These hallucinations compromise trust and reliability in real-world applications, making their detection and mitigation essential.

\begin{figure} 
    \centering
    \includegraphics[width=1.0\columnwidth]{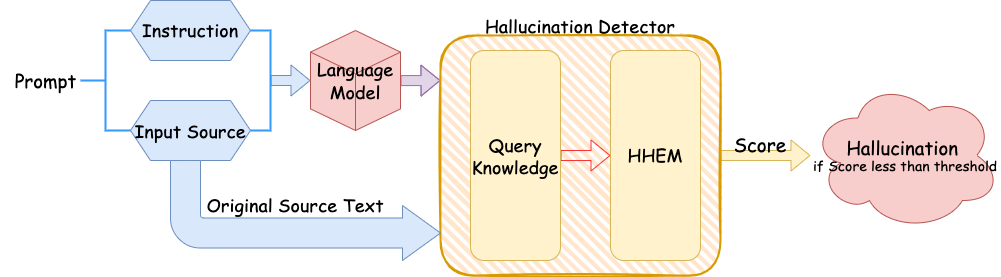}
    \caption{Our work is structured around a comprehensive pipeline designed to identify and rectify hallucinations through a multi-stage factual checking process. The hallucinations are generated by the language model based on the given prompt, which consists of an input source and an instruction. The hallucination detector evaluates the generated response by querying external knowledge and applying the HHEM method to compute a hallucination score. If the score falls below a certain threshold, the response is classified as a hallucination.}
    \label{fig:design}
\end{figure}

To address this, we propose a structured hallucination detection framework that systematically evaluates model-generated text through query decomposition, knowledge retrieval, and factual consistency analysis. Our approach leverages the Hughes Hallucination Evaluation Model (HHEM) to compute a hallucination score, determining whether a response aligns with external knowledge sources. By integrating these verification mechanisms, we aim to enhance LLM reliability, factual accuracy, and robustness. Figure~\ref{fig:design} illustrates our pipeline, which is detailed in the following sections.
\paragraph{Query Decomposition and Knowledge Retrieval}
To improve the accuracy of factual validation, we decompose user queries into structured sub-queries, enabling precise retrieval of relevant external knowledge. This decomposition-based approach ensures that each query focuses on a single logical step, reducing ambiguity in multi-hop retrieval.

\paragraph{Knowledge Optimization}
Raw retrieved knowledge often contains redundant or verbose content, which can obscure factual validation. To enhance consistency, we refine retrieved information into structured and unstructured representations. Structured knowledge is represented as triplets for logical reasoning, while unstructured knowledge provides concise summaries of retrieved text, improving model interpretability.

\paragraph{Hallucination Evaluation using HHEM}
We assess the factuality of generated responses using the Hughes Hallucination Evaluation Model (HHEM), which computes a hallucination score by comparing model-generated text with retrieved knowledge. This score determines factual consistency, enabling classification of responses as hallucinated or reliable. The evaluation is further quantified using standard classification metrics, including F1-score, True Positive Rate (TPR), and True Negative Rate (TNR), ensuring robust assessment of hallucination detection performance.

\subsection{Query Decomposition and Knowledge Retrieval}

Ensuring the verifiability of model-generated responses remains a critical challenge in language model research. Traditional retrieval systems often struggle with complex, multi-hop queries, leading to incomplete or imprecise evidence retrieval. To address this, we introduce a decomposition-based retrieval framework that enhances factual verification by iteratively structuring queries and optimizing knowledge retrieval.

\subsection{Query Decomposition}

To improve retrieval accuracy, we decompose complex user queries into a sequence of structured sub-queries, facilitating one-hop retrieval instead of ambiguous multi-hop queries. Inspired by ReAct\cite{yao2023reactsynergizingreasoningacting}, this method iteratively interacts with external sources, retrieving and verifying factual evidence step by step.

\paragraph{Granularity of Queries}  
Decomposing multi-hop queries into atomic logical queries enhances retrieval precision by ensuring that each step focuses on a single logical operation. This approach minimizes ambiguity and improves the quality of retrieved evidence.

\paragraph{Query Formulation}  
We employ a dual-query strategy to optimize factual retrieval. General queries are broad and avoid speculative details, such as \textit{“Who composed the score for Star Wars?”}, ensuring the retrieval of comprehensive background knowledge. Specific queries explicitly validate model-generated claims, such as \textit{“Did Joy Williams compose the score for Star Wars?”}, focusing on precise fact-checking.  

By iteratively retrieving and verifying evidence at each step, this approach enhances retrieval robustness and mitigates hallucinations.

\subsection{Knowledge Retrieval and Optimization}

Following query decomposition, sub-queries are executed against external knowledge sources using Retrieval-Augmented Generation (RAG) frameworks.

\paragraph{Knowledge Sources}  
Our approach incorporates Wikipedia-based structured databases for authoritative information, along with dense retrieval models such as ColBERT v2 \cite{santhanam-etal-2022-colbertv2} and PLAID \cite{PLAID} to improve retrieval efficiency and accuracy.

\paragraph{Knowledge Optimization}  
Raw retrieved content often contains redundant or noisy information, necessitating refinement for improved factual validation. We categorize knowledge representations into two primary forms:  

Unstructured knowledge extracts concise summaries from retrieved text, such as \textit{“Star Wars, released in 1977, is a space-themed film introducing Luke Skywalker.”}, providing contextually relevant details.  

Structured knowledge encodes facts as triplet relations, such as \textit{(Star Wars, was released in, 1977)}, facilitating symbolic reasoning and structured verification.

This multi-representational approach facilitates fact-based reasoning, improving both retrieval accuracy and hallucination detection, ultimately enhancing the reliability of language model outputs.

\subsection{Hallucination Detection Using HHEM}
To assess the factuality of the generated response, we introduce the Hughes Hallucination Evaluation Model (HHEM), which quantifies hallucinations based on factual grounding. The HHEM model assigns a hallucination score by computing the similarity between the generated response and retrieved knowledge.

\paragraph{Hallucination Score Calculation}  
The hallucination score is computed as follows:

\begin{equation}
    S_h = f(G, K)
\end{equation}

where \( G \) represents the generated response, \( K \) represents the retrieved knowledge, and \( f(G, K) \) is a factual consistency function that measures semantic similarity and contradiction detection. A lower hallucination score (\( S_h \)) indicates a higher likelihood of hallucination.

\subsubsection{Final Judgment and Evaluation Metrics}
To classify hallucinations, we compare \( S_h \) against a predefined threshold \( \tau \):

\paragraph{Classification Rule}  
If \( S_h < \tau \), the response is labeled as hallucinated. Otherwise, it is considered factually reliable.

\paragraph{Evaluation Metrics}  
We evaluate hallucination classification using three key metrics.

True Positive Rate (TPR) measures the proportion of actual hallucinations correctly identified.

\begin{equation}
    \text{TPR} = \frac{\text{True Positives}}{\text{True Positives} + \text{False Negatives}}
\end{equation}

True Negative Rate (TNR) measures the proportion of non-hallucinated responses correctly classified.

\begin{equation}
    \text{TNR} = \frac{\text{True Negatives}}{\text{True Negatives} + \text{False Positives}}
\end{equation}

F1-Score quantifies the balance between precision and recall in detecting hallucinated responses.

\begin{equation}
    F_1 = \frac{TPR + TNR}{2}
\end{equation}

By optimizing the hallucination score threshold \( \tau \), we aim to maximize the F1-score while maintaining a high TPR and TNR, ensuring robust hallucination detection.

\section{Experiments}

\subsection{Implementation and Environment Setup}
We implement our algorithm using Python and set up the environment by cloning the KnowHalu repository from GitHub. A virtual environment is created using `virtualenv`, followed by the installation of required dependencies. Additionally, we download the necessary language models from Hugging Face, as mentioned in the previous sections. To enhance hallucination evaluation, we replace the original judgment model in KnowHalu with our proposed Hughes Hallucination Evaluation Model (HHEM). This modification allows us to systematically evaluate hallucinations based on query decomposition and retrieved knowledge.

\subsection{Experiments Result}
\subsubsection{Hallucination Detection on QA Task}
\begin{figure}[t] 
    \centering
    \includegraphics[width=1.0\columnwidth]{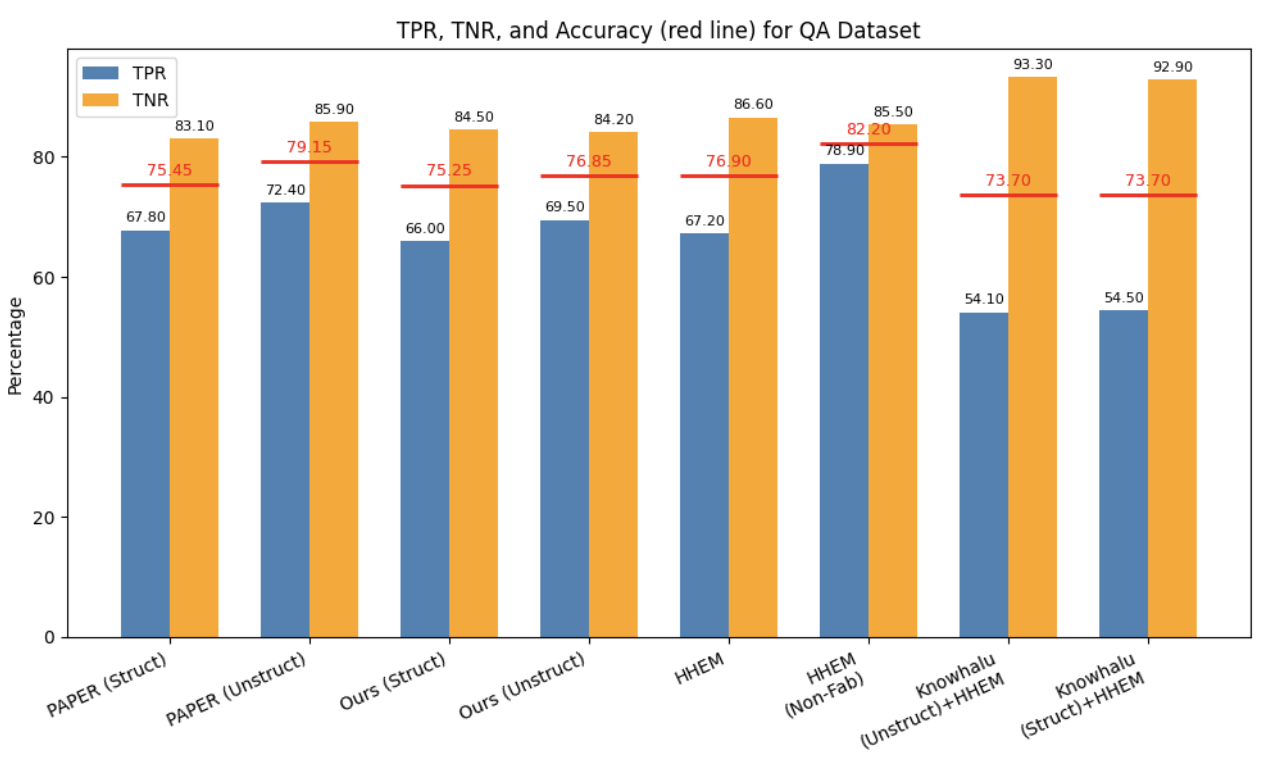} 
    \caption{Results of QA dataset-Starling-LM-7B-alpha}
    \label{fig:qa}
\end{figure}

In this section, we present the results of hallucination detection on the HaluEval QA dataset, comparing multiple methods, including the original PAPER model, our replication, and the HHEM framework. Our replication closely matched the performance of the PAPER model, with minor variations in TPR, TNR, and accuracy due to a necessary reduction in the test set size from 10,000 to 1,000 samples. This reduction was required due to computational constraints, which slightly impacted the dataset distribution. Despite this, the consistency of performance across models demonstrates the robustness and reproducibility of the hallucination detection methodology. Table \ref{table:qa-accuracies} presents a detailed comparison of the methods. The HHEM framework emerged as a significantly more efficient alternative to KnowHalu while maintaining comparable or superior detection accuracy. Notably, HHEM completed hallucination detection in just 10 minutes, compared to approximately 8 hours required by KnowHalu. Despite this drastic reduction in processing time, HHEM achieved a TPR of 67.2\%, a TNR of 86.6\%, and an overall accuracy of 76.9\%. The visualization in Figure\ref{fig:qa} further illustrates this comparison.

\begin{table}[t]
\caption{Results of QA dataset-Starling-LM-7B-alpha}
\label{table:qa-accuracies}
\centering
\resizebox{\columnwidth}{!}{ 
\begin{tabular}{lcccc}
    \toprule
    Method & TPR & TNR & Accuracy & Time \\
    \midrule
    PAPER (Structured)                & 67.80 & 83.10 & 75.45 & - \\
    PAPER (Unstructured)              & 72.40 & 85.90 & 79.15 & - \\
    Ours (Structured)                 & 66.00 & 84.50 & 75.25 & 8h query and 3h judge \\
    Ours (Unstructured)               & 69.50 & 84.20 & 76.85 & 8h query and 3h judge \\
    HHEM                              & 67.20 & 86.60 & 76.90 & 10 min judge \\
    HHEM (Non-Fabrication Checking)   & 78.90 & 85.50 & 82.20 & 1h judge \\
    Knowhalu (Unstructured) + HHEM    & 54.10 & 93.30 & 73.70 & 8h query and 10 min judge \\
    Knowhalu (Structured) + HHEM      & 54.50 & 92.90 & 73.70 & 8h query and 10 min judge \\
    \bottomrule
\end{tabular}
}
\end{table}

The improved version of HHEM, which integrates non-fabrication checking, achieved the highest TPR of 78.9\% and an accuracy of 82.2\%. This approach builds on the method introduced in the KnowHalu paper, verifying facts before making predictions. The additional verification step enhances recall by enabling the model to correctly flag more hallucinations while maintaining accuracy. Despite requiring only one additional hour of processing time, HHEM with non-fabrication checking significantly outperformed standard HHEM and KnowHalu, as depicted in Figure \ref{fig:qa}.

Our results suggest that HHEM is a highly efficient and accurate hallucination detection method. The substantial reduction in inference time compared to KnowHalu makes it a practical solution for real-world applications without compromising detection performance. The addition of non-fabrication checking further boosts effectiveness, particularly for datasets with relatively short knowledge-answer pairs, where factual verification is more straightforward. This finding underscores the potential of HHEM as a fast, reliable, and computationally efficient framework for hallucination detection.

\subsubsection{Hallucination Detection on Summarization Task}
In this section, we present the results of hallucination detection on the HaluEval Summarization dataset, comparing various methods, including the original PAPER model, our replication, and the HHEM framework. Unlike the QA task, the summarization task posed a different challenge due to the inherent nature of summarization, where hallucinations are often localized within longer passages. This led to significant variations in performance across different models.

Table \ref{table:summarization-results} presents a detailed comparison of the methods. The results indicate that KnowHalu performed the best in this task, while HHEM alone struggled to achieve high detection accuracy. The primary reason for this discrepancy is that, in QA tasks, both the original text and answers are typically short, making it relatively straightforward for HHEM to verify factual consistency. However, in summarization tasks, the original text and generated summaries are much longer, often containing a mix of factual and hallucinated information within a single passage. This makes hallucination detection inherently more complex.

\begin{table}[H]
\caption{Results of Summarization dataset-Starling-LM-7B-alpha}
\label{table:summarization-results}
\centering
\resizebox{\columnwidth}{!}{ 
\begin{tabular}{lcccc}
    \toprule
    Method & TPR & TNR & Accuracy & Time \\
    \midrule
    PAPER (Structured)                & 80.20 & 45.40 & 62.80 & - \\
    PAPER (Unstructured)              & 65.00 & 67.20 & 66.10 & - \\
    Ours (Structured)                 & 81.00 & 43.40 & 62.20 & 8h query and 3h judge \\
    Ours (Unstructured)               & 68.60 & 62.00 & 65.30 & 8h query and 3h judge \\
    HHEM                              & 32.20 & 79.40 & 55.80 & 10 min judge \\
    Knowhalu (Unstructured) + HHEM    & 54.40 & 69.00 & 61.70 & 8h query and 10 min judge \\
    Knowhalu (Structured) + HHEM      & 53.00 & 68.80 & 60.90 & 8h query and 10 min judge \\
    \bottomrule
\end{tabular}
}
\end{table} 
\begin{figure}[H]
    \centering
    \includegraphics[width=1.0\columnwidth]{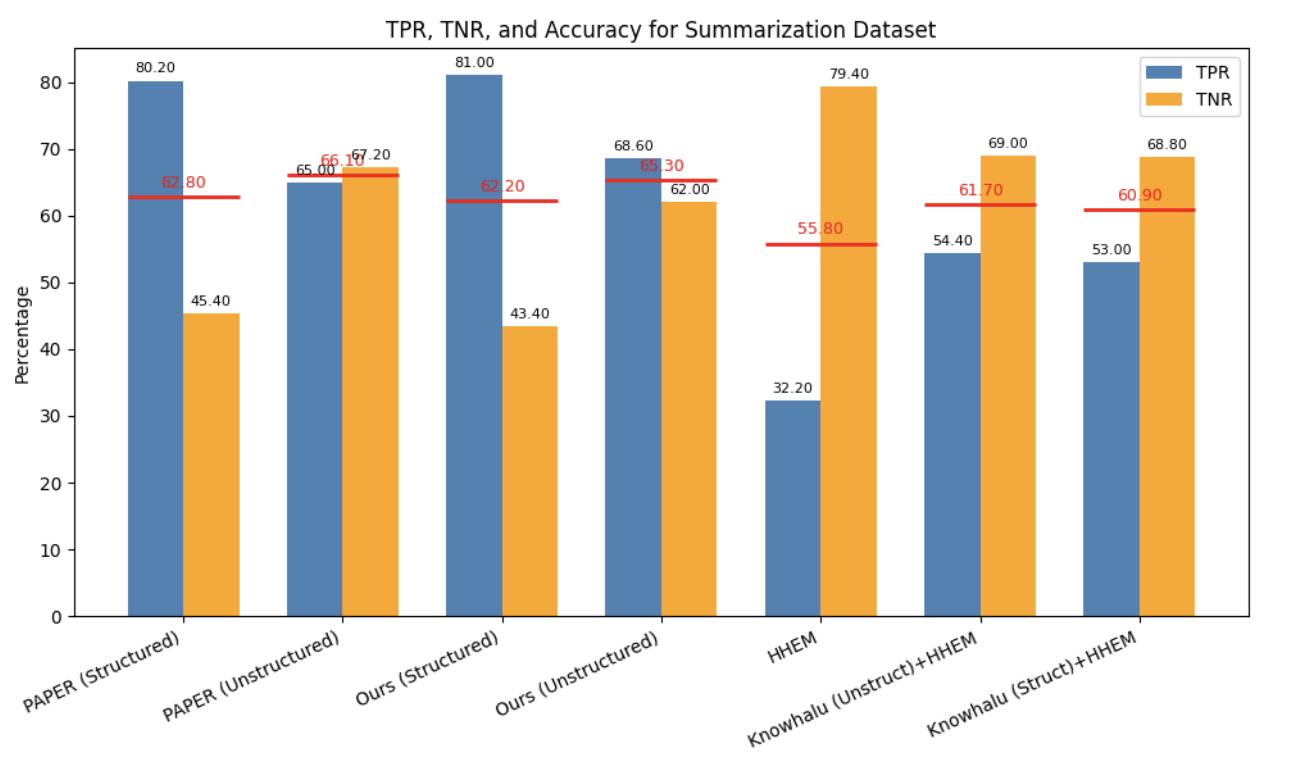}
    \caption{Results of Summarization dataset-Starling-LM-7B-alpha}
    \label{fig:summary}
\end{figure}
Figure \ref{fig:summary} visualizes the comparative performance across different models, highlighting the effectiveness of KnowHalu’s structured approach in improving hallucination detection in summarization tasks.

Unlike in the QA task, where HHEM's simple scoring mechanism performed well, summarization hallucinations required a more refined approach. The standard HHEM model treats the entire summary as a single unit, calculating one global score. Since most sentences within a summary are factually correct, the overall score remains high, leading to a low TPR. This resulted in the model failing to detect hallucinations effectively, as it tended to classify most summaries as non-hallucinated. 

To address this challenge, we incorporated KnowHalu's chain-of-thought reasoning and retrieval method into HHEM. This improved method breaks down the summary into smaller segments and retrieves corresponding knowledge directly from the original text. If any segment of the summary scores low on factual consistency, we consider the entire summary as hallucinated and adjust the HHEM score accordingly. This segmentation-based approach significantly improved hallucination detection by allowing the model to flag smaller hallucinated parts rather than evaluating the entire summary as a whole.

Our results show that using segmentation-based verification is crucial for hallucination detection in summarization tasks. By introducing this method, we observed a clear improvement in TPR, particularly at a threshold of 0.5, where the model was able to detect significantly more hallucinated summaries. These improvements provide greater confidence in leaderboard results and demonstrate the importance of tailored evaluation methods for different NLP tasks.

\subsubsection{Different Model Results of Summarization Task}
In this section, we explore two key aspects of summarization performance across different large language models (LLMs). First, we analyze the length distribution of generated summaries using a boxplot visualization. Then, we evaluate the models' hallucination tendencies based on HHEM scores and present leaderboard results.

\begin{table}[t]
\caption{Results of Summarization Hallucinations Detection}
\label{table:summarization-hallucination}
\centering
\resizebox{\columnwidth}{!}{ 
\begin{tabular}{lcc}
    \toprule
    Method & Accuracy & Hallucination Score \\
    \midrule
    gemma-7b-it                        & 0.75 & 0.21 \\
    Llama-3.1-8B-Instruct              & 0.72 & 0.25 \\
    internlm3-8b-instruct              & 0.69 & 0.26 \\
    Qwen2.5-7B-Instruct                & 0.68 & 0.27 \\
    Llama-2-7b-chat-hf                 & 0.66 & 0.26 \\
    DeepSeek-R1-Distill-Llama-8B       & 0.65 & 0.30 \\
    Mistral-7B-Instruct-v0.3           & 0.65 & 0.27 \\
    Qwen2.5-3B-Instruct                & 0.65 & 0.30 \\
    Qwen2.5-0.5B-Instruct              & 0.55 & 0.36 \\
    Qwen2.5-1.5B-Instruct              & 0.54 & 0.40 \\
    \bottomrule
\end{tabular}
}
\end{table}

\begin{figure}[t]
    \centering
    \includegraphics[width=1.0\columnwidth]{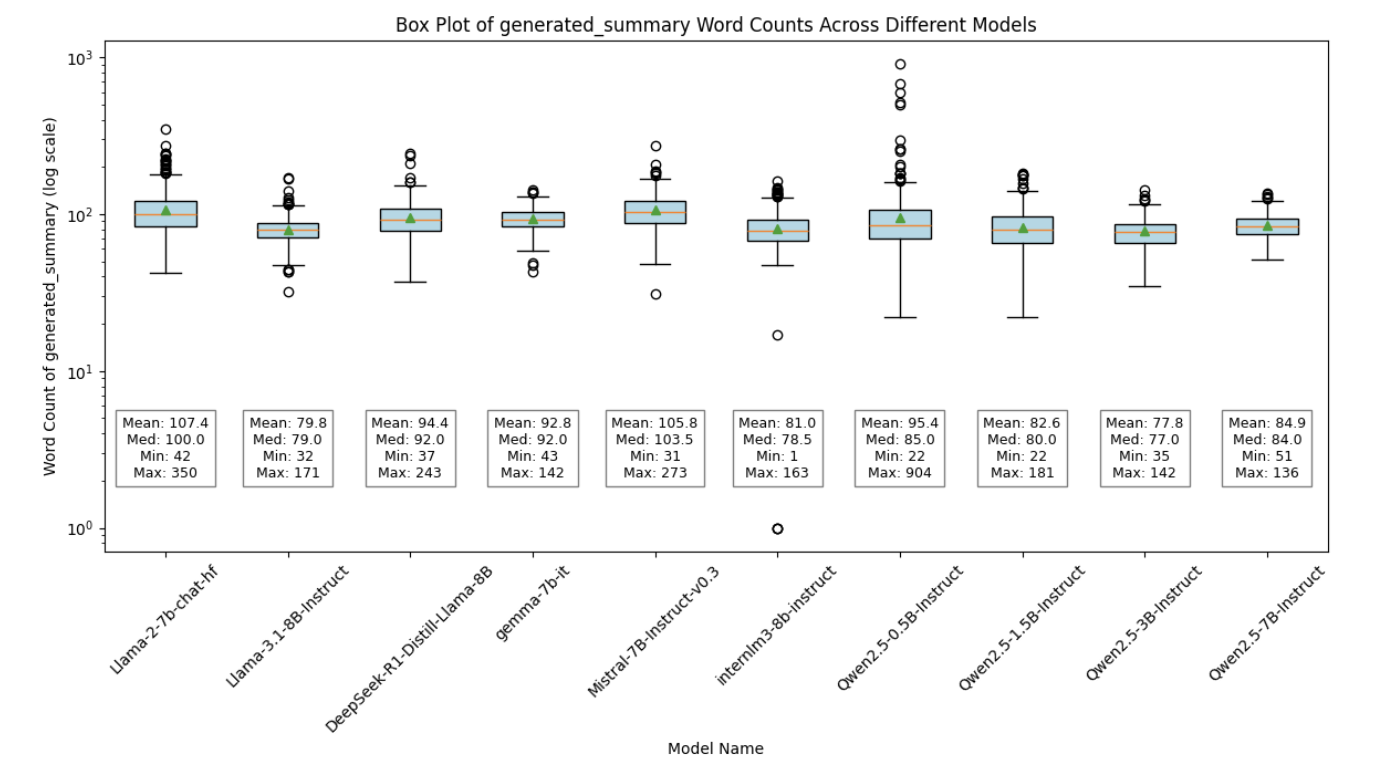}
    \caption{Box Plot of Generated-Summarization Word Counts for Different Models}
    \label{fig:box}
\end{figure}

\paragraph{Analysis of Summarization Lengths}
Figure \ref{fig:box} illustrates the distribution of generated summary lengths across different models. Some models exhibit extreme outliers, with summaries as short as a single word or as long as 904 words. These inconsistencies suggest that certain models fail to adhere to prompt constraints, resulting in outputs that deviate significantly from expected lengths.

A clear trend emerges among the Qwen2.5 series models. As the parameter size increases, these models produce more consistent summaries, with word counts clustering around the 100-word limit. This indicates that larger models are better at following length constraints, likely due to improved comprehension of task instructions. Smaller models, in contrast, generate more variable summaries, leading to increased occurrences of excessively short or long outputs.

\paragraph{Leaderboard Results and Hallucination Performance}

Table \ref{table:summarization-hallucination} presents the accuracy and hallucination scores for different LLMs. We measure accuracy as the proportion of summaries with HHEM scores above 0.5, while the hallucination score is calculated as one minus the average HHEM score.

Our results indicate that gemma-7b-it achieves the highest accuracy of 0.75 with a hallucination score of 0.21, suggesting that it produces the most factually consistent summaries. In contrast, smaller models such as Qwen2.5-0.5B and Qwen2.5-1.5B exhibit higher hallucination scores, indicating a greater tendency to generate misleading or fabricated content.

\subsubsection{CDF of HHEM Scores}
In this section, we analyze the cumulative distribution function (CDF) of HHEM scores for different language models, examining their hallucination tendencies. Additionally, we explore whether increasing the parameter size within the same model series, such as Qwen2.5, results in improved hallucination performance.

\begin{figure}[t]
    \centering
    \includegraphics[width=1.0\textwidth]{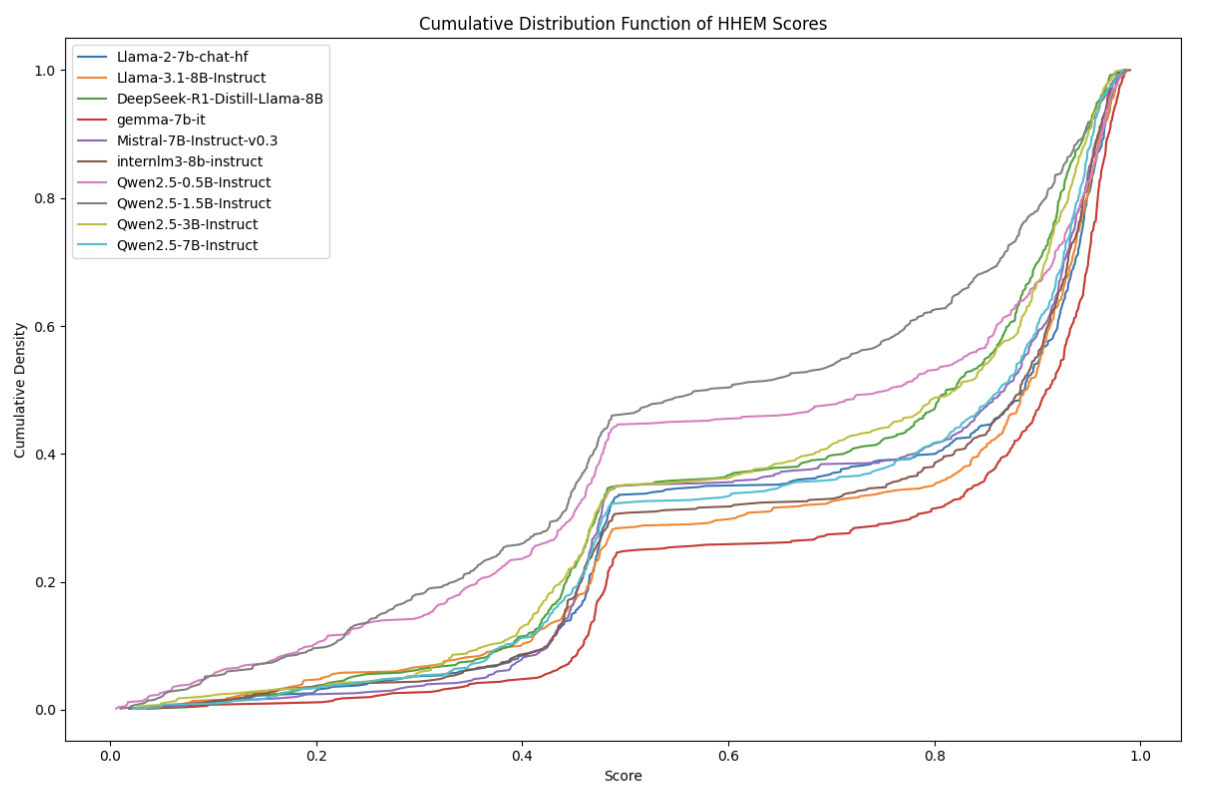}
    \caption{CDF of HHEM Scores Across Models}
    \label{fig:cdf-hhem}
\end{figure}

\begin{figure}[t]
    \centering
    \includegraphics[width=1.0\textwidth]{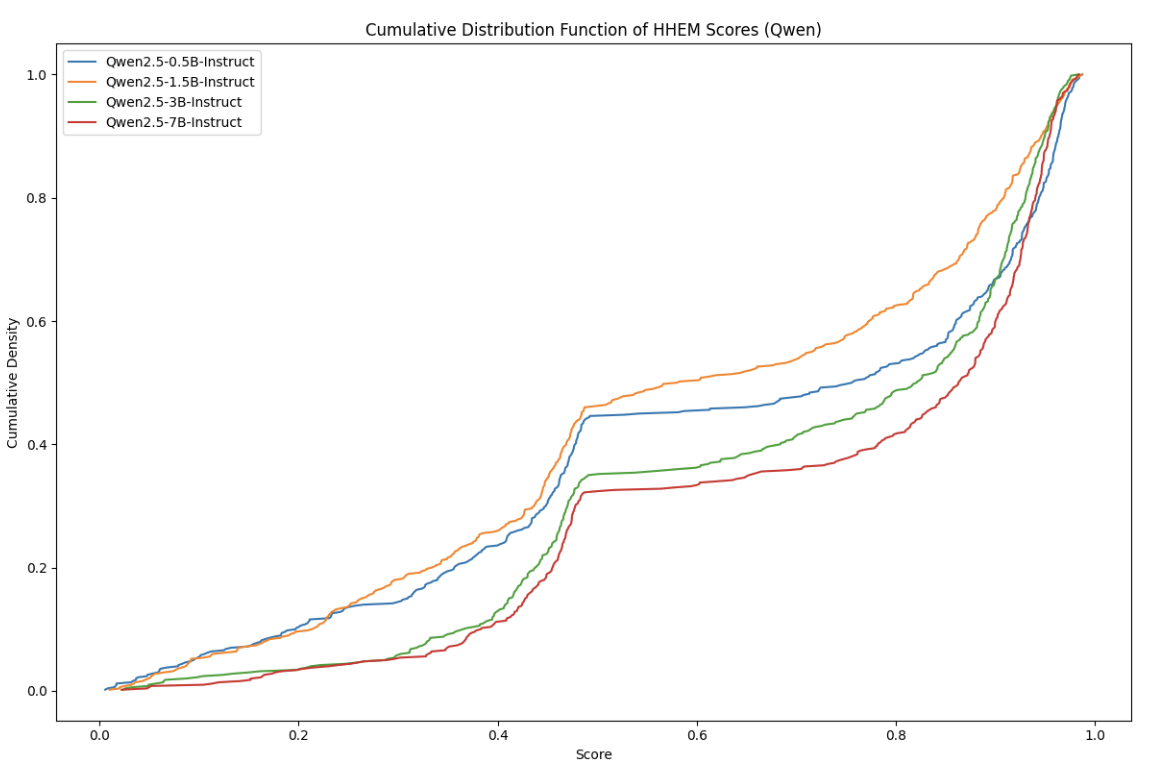}
    \caption{CDF of Qwen Models with Different Parameter Sizes}
    \label{fig:cdf-qwen}
\end{figure}

\paragraph{Analysis of CDF Trends}
Figure \ref{fig:cdf-hhem} illustrates the CDF of HHEM scores across multiple language models. The distribution highlights a clear turning point where hallucinated summaries diverge from factually accurate ones. The lower right section of the graph represents high-quality summaries with minimal hallucinations, while the upper left section indicates a higher prevalence of hallucinated content. A well-performing model should have a steeper curve approaching 1.0, signifying better factual reliability.

A key observation is that models with larger parameter sizes tend to exhibit lower hallucination rates, reflecting more stable and coherent outputs. This is particularly evident in models with seven to nine billion parameters, which consistently outperform smaller models.

\paragraph{Impact of Parameter Scaling in Qwen Models}
Figure \ref{fig:cdf-qwen} focuses on the Qwen2.5 model series, comparing different parameter sizes, including 0.5B, 1.5B, and 3B. Interestingly, while increasing model size generally reduces hallucinations, the trend is not strictly linear. Specifically, Qwen2.5-1.5B exhibits higher hallucination variability compared to both its smaller (0.5B) and larger (3B) counterparts. This suggests that intermediate-sized models may have more unstable factual grounding, potentially due to incomplete knowledge retention during pretraining.

\paragraph{Insights and Implications}
Our findings indicate that while larger models generally perform better in reducing hallucinations, smaller models can sometimes exhibit unpredictable variations. This suggests that model size alone is not a definitive predictor of hallucination robustness. Instead, factors such as pretraining data quality, instruction tuning, and retrieval augmentation may play crucial roles in mitigating hallucinated content.

These insights emphasize the importance of evaluating language models holistically, rather than relying solely on parameter count as a performance metric. Future research should explore how knowledge retrieval mechanisms and fine-tuning strategies can further enhance factual consistency across different model scales.

\section{Discussion}

In our summarization dataset experiment, we evaluated key metrics and recorded the processing time for each method. Notably, the summarization task yielded results that differ substantially from our observations in the QA experiments. Specifically, while HHEM alone performed poorly on summarization, Knowhalu emerged as the top-performing method.

This discrepancy can be attributed to the inherent differences between summarization and QA tasks. In QA, the original text and the generated answer are typically concise, making it relatively straightforward for HHEM to determine whether a knowledge–answer pair is correct. In contrast, summarization involves much longer texts for both the source and the generated summary. Consequently, only a few sentences in an extensive summary may hallucinate, while the majority remain factually accurate.

The traditional HHEM approach treats the entire summary as a single unit and computes one overall score. Because most sentences in the summary are correct, the aggregated HHEM score tends to be high, which results in a low TPR as the model often fails to detect the few hallucinated segments. Essentially, this causes the model to classify nearly every summary as non-hallucinated despite the presence of some inaccuracies.

To address this limitation, we integrated the Knowhalu chain-of-thought and retrieval method into our framework. Our approach involves partitioning the summary into smaller segments and retrieving corresponding knowledge directly from the original text for each segment. If any segment produces a low knowledge-summary match score, we treat the entire summary as hallucinated and accordingly reduce the overall HHEM score by half. Our experiments revealed a clear turning point at a score of 0.5, which significantly enhances the TPR, allowing us to detect a greater number of hallucinated summaries. With this improved detection capability, we are in the process of rebuilding our leaderboard with higher confidence in the evaluation results.

\section{Limitation and future work}
Although our framework achieved better performance and significantly reduced computation time on the Q\&A dataset, the KnowHalu+HHEM framework did not outperform the original KnowHalu. Therefore, we need to explore methods to improve HHEM’s accuracy—particularly for the summarization task, as the HHEM leaderboard uses a summarization format instead of Q\&A—or consider alternative frameworks that match KnowHalu's performance while preserving computational efficiency. Moreover, although our framework successfully reduces judgment time, the query phase remains lengthy—taking roughly eight hours—since it still relies on the same chain-of-thought and retrieval structure. Future research should focus on reducing query time while balancing the trade-off between accuracy and computational efficiency.In addition, a promising direction for future work involves the integration of reinforcement learning (RL) to detect and mitigate hallucinations. Recent studies have demonstrated that reinforcement learning from human feedback (RLHF) can fine-tune language models by providing reward signals that encourage the generation of factually correct and contextually relevant responses.
By extending our work in above potential directions, we anticipate a more robust system capable of reducing both hallucination frequency and query latency while maintaining high performance across different tasks.

\section{Acknowledgments}

I would like to thank Professor Yuan Tian and Ying Xu, their valuable suggestions and great support throughout this project. This note was not reviewed, approved, or endorsed by Prof. Yuan Tian. Any errors are solely the author's responsibility. I would like to thank John Liu for his valuable contribution.

\clearpage
\bibliography{arxiv}
\bibliographystyle{ims}

\end{document}